\documentclass[a4paper]{article}
\usepackage{iwslt18,amssymb,amsmath,epsfig}
\usepackage{graphicx}

\usepackage{amsmath}

\usepackage{cite}

\usepackage{caption}
\usepackage{amsfonts}
\usepackage{multirow}

\usepackage{wasysym}

\usepackage[T1]{fontenc}

\usepackage{tabularx}
\def\reg{{\rm\ooalign{\hfil
     \raise.07ex\hbox{\scriptsize R}\hfil\crcr\mathhexbox20D}}}

\title{Breaking the Data Barrier: Towards Robust Speech Translation via Adversarial Stability Training}

\makeatletter
\def\name#1{\gdef\@name{#1\\}}
\makeatother
\name{{\em Qiao Cheng, Meiyuan Fang, Yaqian Han,}\\
      {\em Jin Huang, Yitao Duan}}

\address{NetEase Youdao Information Technology (Beijing) Co., LTD. \\
Beijing, China\\
{\small \tt \{chengqiao, fangmeiyuan, hanyaqian, huangjin, duan\}@rd.netease.com}
}
\begin{document}
\maketitle
\begin{abstract}
In a pipeline speech translation system, automatic speech recognition (ASR) system will transmit errors in recognition to the downstream machine translation (MT) system. A standard machine translation system is usually trained on parallel corpus composed of clean text and will perform poorly on text with recognition noise, a gap well known in speech translation community. In this paper, we propose a training architecture which aims at making a neural machine translation model more robust against speech recognition errors. Our approach addresses the encoder and the decoder simultaneously using adversarial learning and data augmentation, respectively. Experimental results on IWSLT2018 speech translation task show that our approach can bridge the gap between the ASR output and the MT input, outperforms the baseline by up to 2.83 BLEU on noisy ASR output, while maintaining close performance on clean text.
\end{abstract}

\section{Introduction}

A pipeline spoken language translation (SLT) system integrates two main modules: source language speech recognition (ASR) and source-to-target text translation (MT). The ASR system transforms the input audios into automatic transcriptions which the MT system translates into texts in the target language. However, since the two systems are trained separately using different parallel corpora, there is a mismatch between the output of ASR and the expected input of MT system. The former does not provide punctuation or case information, and contains recognition errors such as omission, repetition, homophones errors, etc. In contrast, a standard MT system is usually trained on written text which is much cleaner.

Although an end-to-end speech translation model \cite{weiss2017sequence, berard2016listen} seems to be a promising solution to alleviate the problem of error propagation, due to the scarcity of training data in the form of speech aligned with text and translation, currently the pipeline system is still the best performing model so far. 

Prior works attempting to enhance the robustness of speech translation suffer from two main limitations: they either require training data in the form of speech aligned with either translation or both transcription and translation, or use artificially generated noise. The first problem is faced by end-to-end models \cite{weiss2017sequence, berard2016listen} and \cite{peitz2012spoken,sperber2017neural}. Such speech-to-transcription-to-translation data is extremely costly to acquire since it requires two human (even expert) labeling processes for each sentence. We are only aware of four such corpora that are publicly available:  Fisher and Callhome Spanish-English Speech Translation Corpus \cite{post2013improved}, Speech translation TED corpus \cite{jan2018iwslt}, MaSS \cite{boito2019mass} and MuST-C \cite{di2019must}. All of them are \emph{far} less than what is necessary for training even a single ASR or MT model. The works relying on artificial noise include \cite{li2018improving, sperber2017toward} etc. They only achieve minor improvements on noisy input but harm the translation quality on clean text. 

On the other hand, we observe that, when examined separately, in either ASR or MT, there exist many large parallel corpora, some to the scale of billions of sentences pairs. Yet they cannot be linked together to form the speech-to-translation pairs that the end-to-end models need, or the speech-to-transcription-to-translation triplets that robust pipeline speech translation systems need.

In this paper, we introduce a training architecture that breaks the data barrier, allowing us to use both ASR and MT corpora simultaneously without having to explicitly linking the two. The main idea is to force both the encoder and decoder of a neural machine translation (NMT) model to behave consistently in face of ASR-induced noise. The  speech-to-transcription data is introduced into NMT's training process to provide additional supervise signals for both the encoder and decoder. Those additional supervise signals do not rely on the existence of translation so they are decoupled from NMT's usual objective. Therefore the two types of data can be used independently. Our work has the following advantages compared with previous works:

\begin{itemize}
\item \textbf{Actual ASR Adaptation.} Instead of simulated noise, our architecture can incorporate real ASR output, significantly improving the MT system's performance on actual speech translation tasks.

\item \textbf{Easy Data Acquisition.}  Ours is the \emph{first} unified training framework that utilizes both actual ASR and NMT corpora but does \emph{not} require speech-to-transcription-to-translation alignments. Instead, datasets for ASR and NMT can be collected independently and our network combines them in a single training process. This is a huge advantage since parallel corpora for both ASR and NMT are relatively easy to obtain separately but it is very costly to bridge the gap between the two to produce a sufficiently large audio-to-text-to-translation dataset. 

\item\textbf{Robustness.} Due to its ability to use both ASR and NMT corpora simultaneously, our approach improves the performance on  noisy input without sacrificing the translation quality on clean data.
\end{itemize}

The paper is organized as follows. In the next section, we give a brief overview of related work. In Section 3, we describe our MT architecture. Finally, we discuss the experimental results in Section 4, followed by a conclusion.

\section{Related work}

Recently some researchers have attempted to address the problem of speech translation using an end-to-end approach \cite{weiss2017sequence, berard2016listen, berard2018end, serdyuk2018towards}. While this is definitely a very promising direction, however, due to the scarcity of speech-to-translation data, these works either use synthetic speech corpora which are generated by a text-to-speech (TTS) system \cite{weiss2017sequence, berard2016listen} or do not directly predict the translation \cite{duong2016attentional, bansal2017towards}. In \cite{duong2016attentional} seq2seq model was used to align speech with translated text, and \cite{bansal2017towards} can only generate bags of words. As demonstrated by the results of IWSLT2018 Speech Translation Task \cite{jan2018iwslt}, at present, the most effective way to do spoken language translation is still a pipeline system. 
 
Sequence-to-sequence models are known to be sensitive to noise \cite{belinkov2017synthetic}. As the translation model has not been trained to cope with ASR errors, they tend to work poorly when stacked with an ASR system \cite{belinkov2017synthetic, le2017disentangling, ruiz2017assessing}. There are a couple of problems. Firstly, the ASR output is devoid of punctuation and case information, which is very important for downstream translation systems. This issue has been addressed by some works such as \cite{cho2017nmt, varavs2018restoring}. Secondly, ASR also introduces other errors such as missing words, repetition, homophones, etc.

In order to solve the homophone noise produced by ASR systems, \cite{tsvetkov2014augmenting} uses a source language  pronunciation dictionary and a language model to simulate possible misidentification errors in ASR and use them to augment the phrase table of an SMT system. \cite{liu2018robust} combined textual and phonetic information in the embedding layer of neural networks. Their approach relies on additional resources and addresses only one type of speech recognition error.

Data augmentation is an important method to improve the performance of neural networks. \cite{li2018improving}  simulates the noise existing in the real output of the ASR system and inject them into the clean parallel data so that NMT can work under similar word distributions during training and testing. \cite{sperber2017toward} shows that with a simple generative noise model, moderate gains can be achieved in translating erroneous speech transcripts, provided that type and amount of noise are properly calibrated. \cite{cheng2018towards} uses generic Gaussian noise to perturb the inputs to the NMT training and robustify the model through adversarial training. All the above work with simulated noise, which, as will be demonstrated in Section 4, does not realistically reflect the ASR output and causes suboptimal speech translation quality.

Using real ASR output as training input to the translation model is a natural idea. Some work with ASR's final output directly \cite{peitz2012spoken} while some use its internal representations such as word lattices to explicitly model  uncertainty of the upstream system \cite{sperber2017neural}. They all require speech-to-transcript-to-translation data which is very rare.

\section{Approach}

\begin{figure*}[htb]
  \centering
  \includegraphics[scale=.5,width=0.98\textwidth]{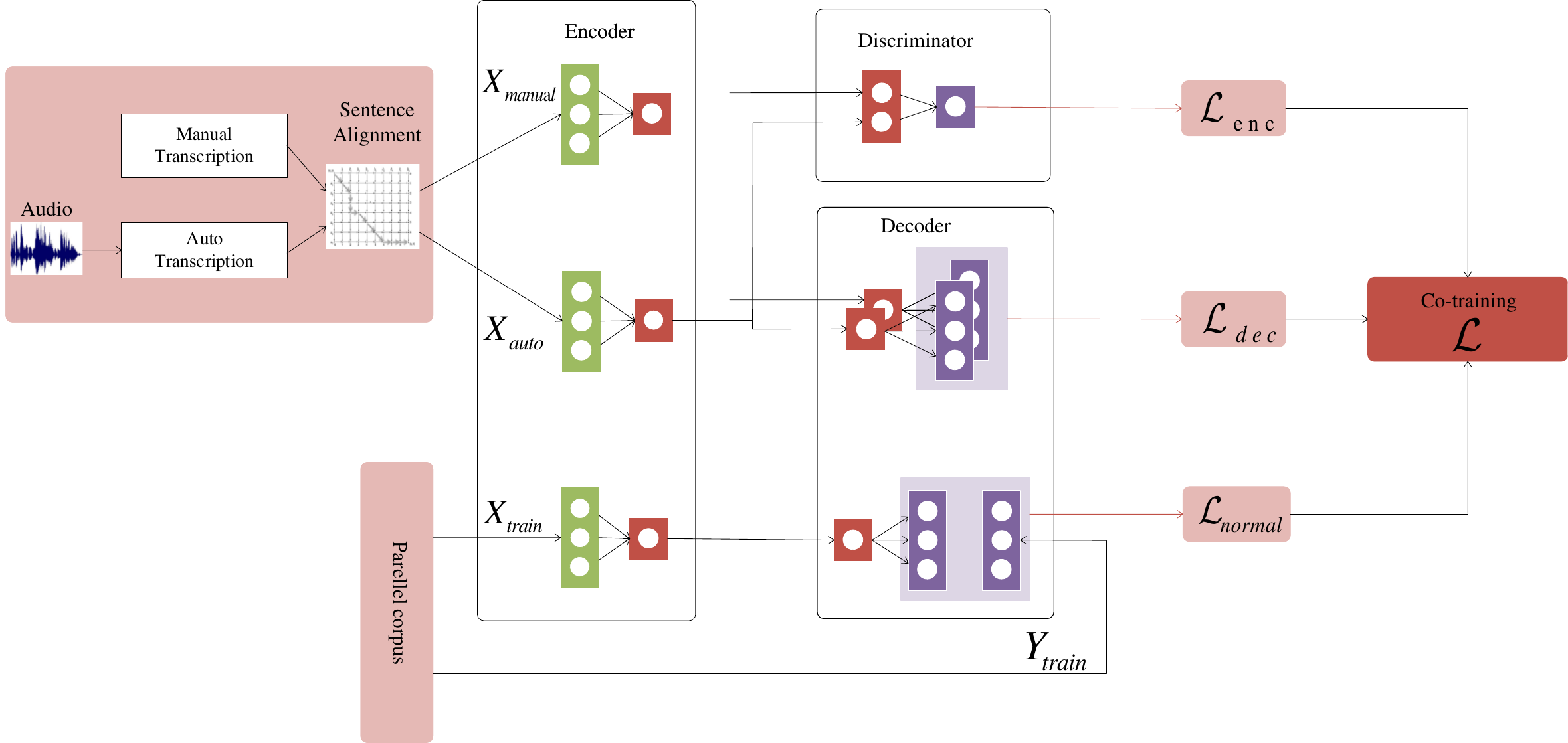}
  \caption{The training architecture. Manual transcription and automatic transcription are aligned before input to the encoder. The encoder output is sent to the discriminator and decoder module. The discriminator forces the encoder to produce similar representations for both clean and noisy versions of the same sentence. The decoder output of manual transcription serves as the reference of automatic transcription. The final training objective is the sum of $\mathcal{L}_{enc}$, $\mathcal{L}_{dec}$, together with the normal objective $\mathcal{L}_{normal}$.}
  \label{fig:speech_production}
\end{figure*}

The proposed training scheme takes two kinds of parallel corpora as input. The first corpus, also called ASR data, consists of speech-transcription pairs. The second corpus is the NMT's training data, in the form of source-translation pairs. 
Our approach consists of two phases: the ASR data processing and the adversarial NMT training. At the first step, we process the ASR data which provides audios and manual transcriptions. A trained ASR model is utilized to recognize the audios, after which a re-segmentation algorithm is used to align the automatic transcriptions with its corresponding manual transcriptions in a sentence-wise manner. The aligned sentence-wise transcription pairs are then sent to the next stage of the proposed training scheme. In the second stage, we treat the manual transcription as a supervised signal for the automatic transcription which is combined with the NMT's usual objectives to co-train the model.

\subsection{Processing the ASR data} 
\label{ssection:processing}

A trained ASR model is used to automatically generate the transcriptions given the audios. Obviously, the proposed approach is highly dependent on the accuracy of the ASR model, and it would be more beneficial to use in-domain data of the given ASR model as well. But to make the work reproducible and consistent among a wide range of ASR data, we use an open source ASR model without limiting the domain of the ASR data. 

In general, the automatic transcriptions generated by an ASR system are not accompanied by their sentence-wise segmentation. However, training an NMT system works with parallel corpora of sentences. Therefore the next step is to align the automatic transcription with the manual transcription and segmentation to generate sentence pairs. We exploit the automatic re-segmentation algorithm described in \cite{matusov2005evaluating} for this purpose.

The re-segmentation algorithm calculates the Levenshtein ratio between automatic and manual transcriptions. By backtracking the decision of the distance algorithm, the alignment of the given word sequence with the existing sentence segmentation and manual transcription can be found for reference. In this way, sentence segmentation is transformed into the recognition of reference. Dynamic programming is used to solve the re-segmentation problem.

\subsection{NMT training architecture}

The conventional NMT model consists of two procedures for projecting a source sentence \textbf{x} to its target sentence \textbf{y}: the encoder \textbf{Enc($\cdot$)} is responsible for encoding \textbf{x} as a sequence of representations $\textbf{H}_\textup{x}$, while the decoder \textbf{Dec($\cdot$)} produces \textbf{y} with $\textbf{H}_\textup{x}$ as input. The model is optimized by minimizing the negative log-likelihood of the target translation \textbf{y} given source sentence \textbf{x}. Denoting the parameters of encoder and decoder as {\boldmath$\theta$}$_{enc}$ and {\boldmath$\theta$}$ _{dec}$, respectively, the loss function $\mathcal{L}$ can be written as 
\begin{equation}
\begin{aligned}
  \mathcal{L}(\textbf{x}, \textbf{y}; \text{\boldmath$\theta$}_{enc}, \text{\boldmath$\theta$}_{dec}) = -\frac{1}{|\textbf{y}|}\sum_{j=1}^{|\textbf{y}|}\log P(y_{j}|\mathbf{y}_{<j},\textbf{x}; \text{\boldmath$\theta$}_{enc}, \text{\boldmath$\theta$}_{dec})
  \label{eqnmt}
\end{aligned}
\end{equation}

Our adversarial training architecture is illustrated in Figure~\ref{fig:speech_production}. As described in the previous subsection, the automatic transcriptions are aligned to the manual transcriptions in sentence level. 
We denote the automatic transcription sentence as $\textbf{x}_\textnormal{auto}$ and the manual transcription sentence as $\textbf{x}_\textnormal{manual}$. 
Note that our main concern is to make the encoder extract similar feature representations from $\textbf{x}_\textnormal{auto}$ and $\textbf{x}_\textnormal{manual}$, and hence obtain similar results on final translation. 
Since the translation process can be viewed as \textbf{y} = \textbf{Dec(\textbf{Enc(\textbf{x}))}}, we introduce two intermediate objectives: $\mathcal{L}_{enc}$ forces the encoder to produce similar representations for $\textbf{x}_\textnormal{auto}$ and $\textbf{x}_\textnormal{manual}$; and  $\mathcal{L}_{dec}$ forces the decoder to produce similar results for $\textbf{H}_\textnormal{auto}$ = \textbf{Enc($\textbf{x}_\textnormal{auto}$)} and $\textbf{H}_\textnormal{manual}$ = \textbf{Enc($\textbf{x}_\textnormal{manual}$)}.

Identical to the work of \cite{cheng2018towards}, $\mathcal{L}_{enc}(\textbf{x}_\textnormal{auto}, \textbf{x}_\textnormal{manual})$ is treated as the training objective of an adversarial learning framework \cite{goodfellow2014generative} . In an adversarial learning framework, there are two networks which we called the generator network and the discriminator network. The output of the generator needs to imitate the real samples in the training set as much as possible. The input of the discriminator is the real sample or the output of the generator and the target of discriminator network is to distinguish the output of the generated network from the real sample while the generator network should cheat the discriminator network as much as possible. The two networks oppose each other and adjust the parameters constantly, so as to make the discriminator network unable to judge whether the output of the generated network is true or not. In this work, the encoder serves as the generator $G$, which defines a policy for generating a sequence of hidden representations of $\textbf{H}_\textup{x}$ given the input sentence \textbf{x}. An additional discriminator $D$ is introduced into the training architecture which distinguishes the encoder output of the automatic transcription $\textbf{H}_\textnormal{auto}$ from the encoder output of the manual transcription $\textbf{H}_\textnormal{manual}$. The goal of the  generator, in this case the encoder is to produce similar output for $\textbf{x}_\textnormal{auto}$ and $\textbf{x}_\textnormal{manual}$ which could fool the discriminator, while the discriminator $D$ tries to correctly distinguish the two outputs.

Formally, the adversarial learning objective can be written as
\begin{equation}
\begin{aligned}
  &\mathcal{L}_{enc} (\textbf{x}_\textnormal{auto}, \textbf{x}_\textnormal{manual};\text{\boldmath$\theta _{enc}$},\text{\boldmath$\theta _{dis}$}) \\
= &\mathbb{E}_{\textbf{x}_\textnormal{manual}\sim S}\left [ -\log D(G(\textbf{x}_\textnormal{manual})) \right ]  \\
+ &\mathbb{E}_{\textbf{x}_\textnormal{auto}\sim S}\left [ -\log (1 - D(G(\textbf{x}_\textnormal{auto}))) \right ] \\
  \label{eq_l1}
\end{aligned}
\end{equation}
where $S$ denotes the set of transcription sentences after alignment. 
Specifically, when backpropagated, the gradients of $\mathcal{L}_{enc}$ are replaced by their additive inverse while other gradients remain unchanged, so that all parameters including encoder parameters {\boldmath$\theta$}$_{enc}$ and the discriminator parameters {\boldmath$\theta$}$_{dis}$ can be updated in tandem. 
In such manner, high efficiency in training is attained.

A common idea to handle $\mathcal{L}_{dec}$ is using the adversarial learning framework\cite{yu2017seqgan}. 
Here we introduce a simpler but effective method. 
We first decode $\textbf{H}_\textnormal{manual}$ and obtain the translation $\hat{\textbf{y}}_\textnormal{manual}$ with regard to $\textbf{x}_\textnormal{manual}$. 
We argue that $\hat{\textbf{y}}_\textnormal{manual}$ is the best quality translation of $\textbf{x}_\textnormal{auto}$ the current NMT could ever produce thus can serve as the reference for $\textbf{x}_\textnormal{auto}$. $\mathcal{L}_{dec}(\textbf{x}_\textnormal{auto}, \hat{\textbf{y}}_\textnormal{manual})$ is then calculated by Eq~\ref{eqnmt} as one training objective. 



Of course, the two loss functions above must be combined with the usual MT loss computed on the translation corpus, $\mathcal{L}_{normal}(\textbf{x}_\textnormal{train}, \textbf{y}_\textnormal{train})$  where $\textbf{x}_\textnormal{train}$ and $\textbf{y}_\textnormal{train}$ denote the parallel translation corpus. $\mathcal{L}_{normal}$ is also calculated by Eq~\ref{eqnmt}. 

Finally, our training objective is the sum of the three:
\begin{equation}
\begin{aligned}
  \mathcal{L} = \alpha \mathcal{L}_{enc} +   \beta \mathcal{L}_{dec} + \mathcal{L}_{normal}
  \label{total_l}
\end{aligned}
\end{equation}
where the two hyper-parameters $\alpha$, $\beta$ balance the weights of different loss functions. 
The impact of the two hyper-parameters will be discussed further in Section 4.2.

The proposed method only alters the NMT's training process, remaining the inference procedure unchanged.

\section{Experiments}

\subsection{Setup}

We conducted experiments on the dataset provided by IWSLT2018 Speech Translation Task, which addresses the problem of translating English audio to German text. The provided training data consists of five parts: TED corpus, Speech-translation TED corpus, TED LIUM corpus, WMT18 data and OpenSubtitles0218. Among them, TED corpus, Speech-translation TED corpus, WMT18 data and OpenSubtitles2018 contain parallel sentence pairs, which are then used to train our baseline NMT model. The statistics of the raw training data set are listed in Table~\ref{tab:text_training_data}. Dev2010 and tst2010 are used as the development set and test set in this experiment respectively. The quality of speech translation is measured by the 4-gram BLEU scores \cite{papineni2002bleu}.

\begin{table}[]
\centering
\caption{Statistics of the Training Data}
\label{tab:text_training_data}
\begin{tabular}{|c|c|}
\hline
Corpus                              & \#Sentences \\ 
\hline
WMT18                             & 41M              \\ 
\hline
OpenSubtitles2018         & 22M              \\ 
\hline
TED WIT3                         & 0.2M             \\ 
\hline
Speech-translation TED & 0.17M            \\ 
\hline
TED LIUM2                      & 0.09M            \\ 
\hline
\end{tabular}
\end{table}

For our baseline model, we followed the normal data preprocessing steps: norm-punc, tokenized and lowercased source and target sides using Moses scripts and we cleaned data by removing sentences whose number of tokens are over 100 and the length ratio of source/target is less than 1/2 or larger than 2. We follow \cite{sennrich2016neural} to split words into subword units. The numbers of merge operations in byte pair encoding (BPE) are set to 64K. Our model uses the Transformer architecture \cite{vaswani2017attention} which is solely based on attention mechanisms and dominates most of the sequence-to-sequence tasks. Our hyper-parameters of the Transformer models follow the transformer\_big configuration of tensor2tensor \cite{vaswani2018tensor2tensor}, an open-source implementation of the Transformer model. Both encoder and decoder have 6 layers. The dimension for hidden layers is 1024 while the inner size of the feed-forward network is 4096. The head number of multi-head attention layer is set to 16. Label smoothing and dropout are adopted in the model training.

Speech-translation TED corpus and TED LIUM corpus provide both audio data and transcriptions. However, since the transcriptions provided in TED LIUM2 are neither human-generated nor human-annotated, we only use the Speech-translation TED corpus in the adversarial training phase. 

IWSLT2018 provides a trained ASR so we can focus on the NMT part of the pipeline. We use the provided ASR system to get the automatic transcriptions from the input audio, and then align these transcriptions with the manual transcriptions following the method presented in Section~\ref{ssection:processing}. The transcription sentences are sorted by Word Error Rate (WER) and the top 1$\permil$ bad sentences are excluded from the subsequent training process.


Our adversarial stability training initializes the model parameters with the baseline model. And the discriminator module optimized by $\mathcal{L}_{enc}$ is composed of two sub-layers. The sentence embedding which is the average of source representations is fed to the discriminator. The first sub-layer of the discriminator is a multi-head self-attention over the output of the encoder stack and the second is a feed-forward network. We tested the effects of $\mathcal{L}_{enc}$ and $\mathcal{L}_{dec}$ on neural machine translation and speech translation respectively. Finally, we applied $\mathcal{L}_{enc}$ and $\mathcal{L}_{dec}$ to our system at the same time and obtained the optimal results.

\begin{figure*}[htb]
\begin{minipage}{0.5\textwidth}
  \centering
  \includegraphics[width=0.85\textwidth]{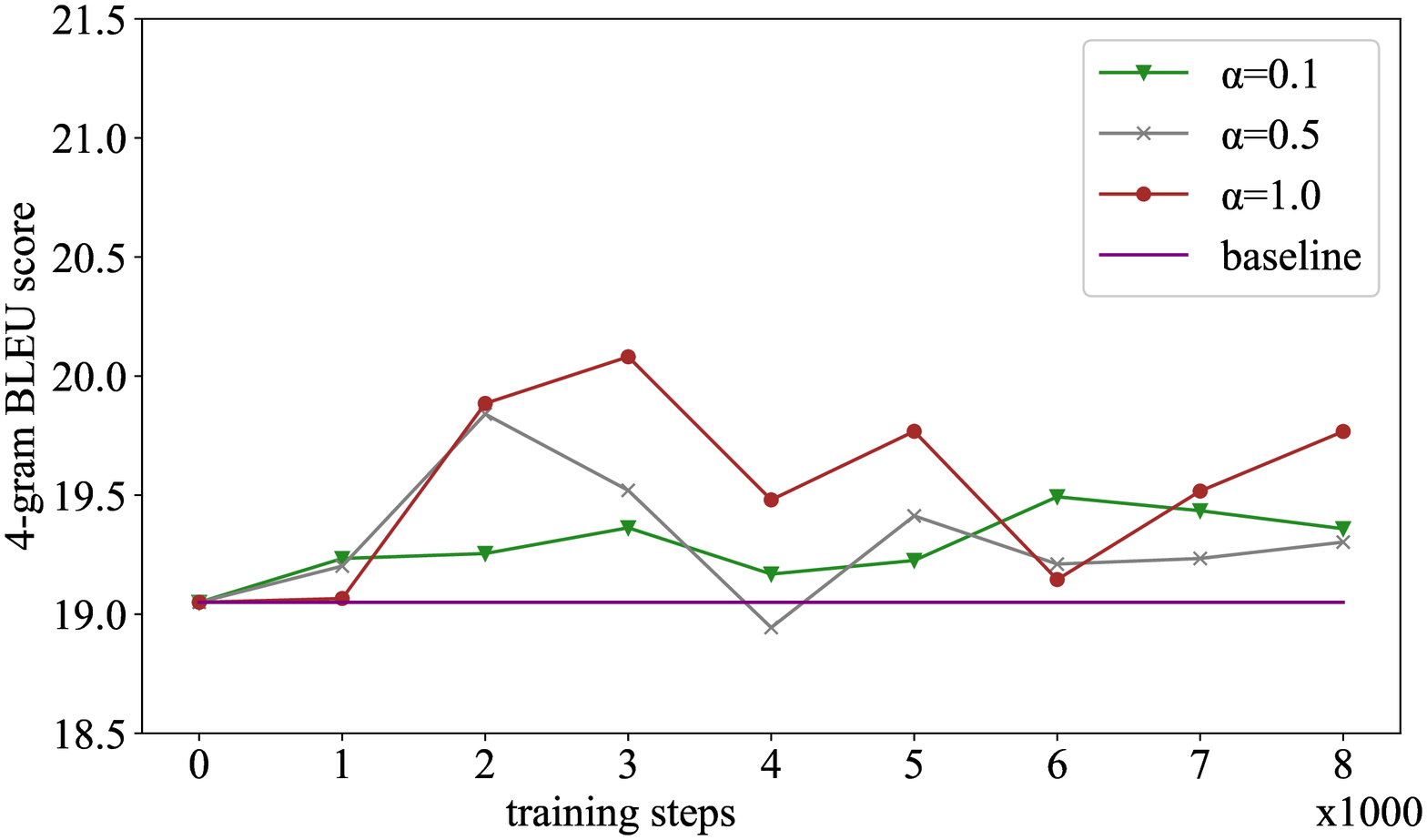}
  \captionsetup{width=.85\linewidth}
  \caption{BLEU scores on tst2010, using ASR transcripts as inputs, varying $\alpha$}
  \label{fig:encoder_asr_alpha}
\end{minipage}\hfill
\begin{minipage}{0.5\textwidth}
  \centering
  \includegraphics[width=0.85\textwidth]{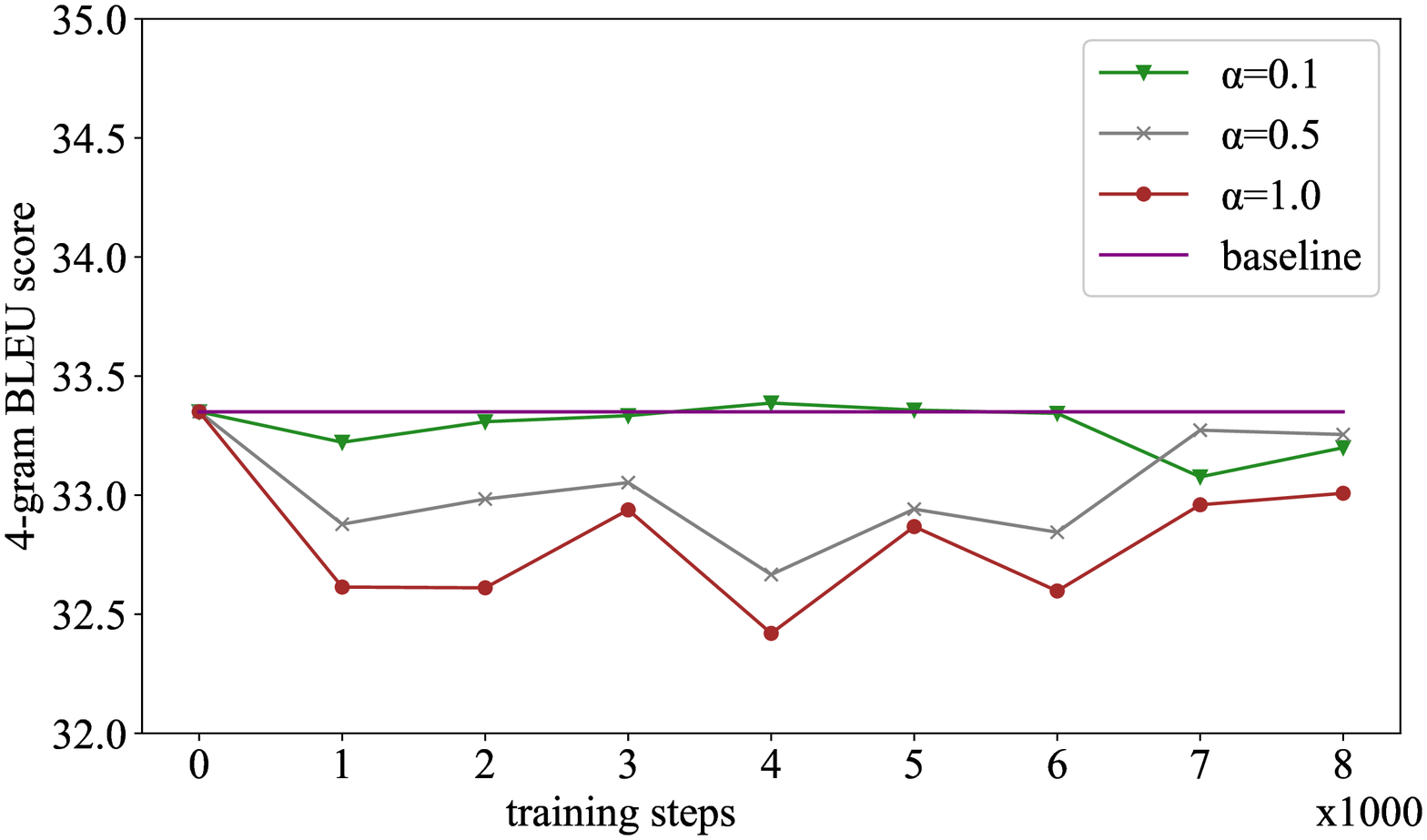}
  \captionsetup{width=.85\linewidth}
  \caption{BLEU scores on tst2010, using clean reference transcripts as inputs, varying $\alpha$}
  \label{fig:encoder_ref_alpha}
\end{minipage}
\begin{minipage}{0.5\textwidth}
  \centering
  \includegraphics[width=0.85\textwidth]{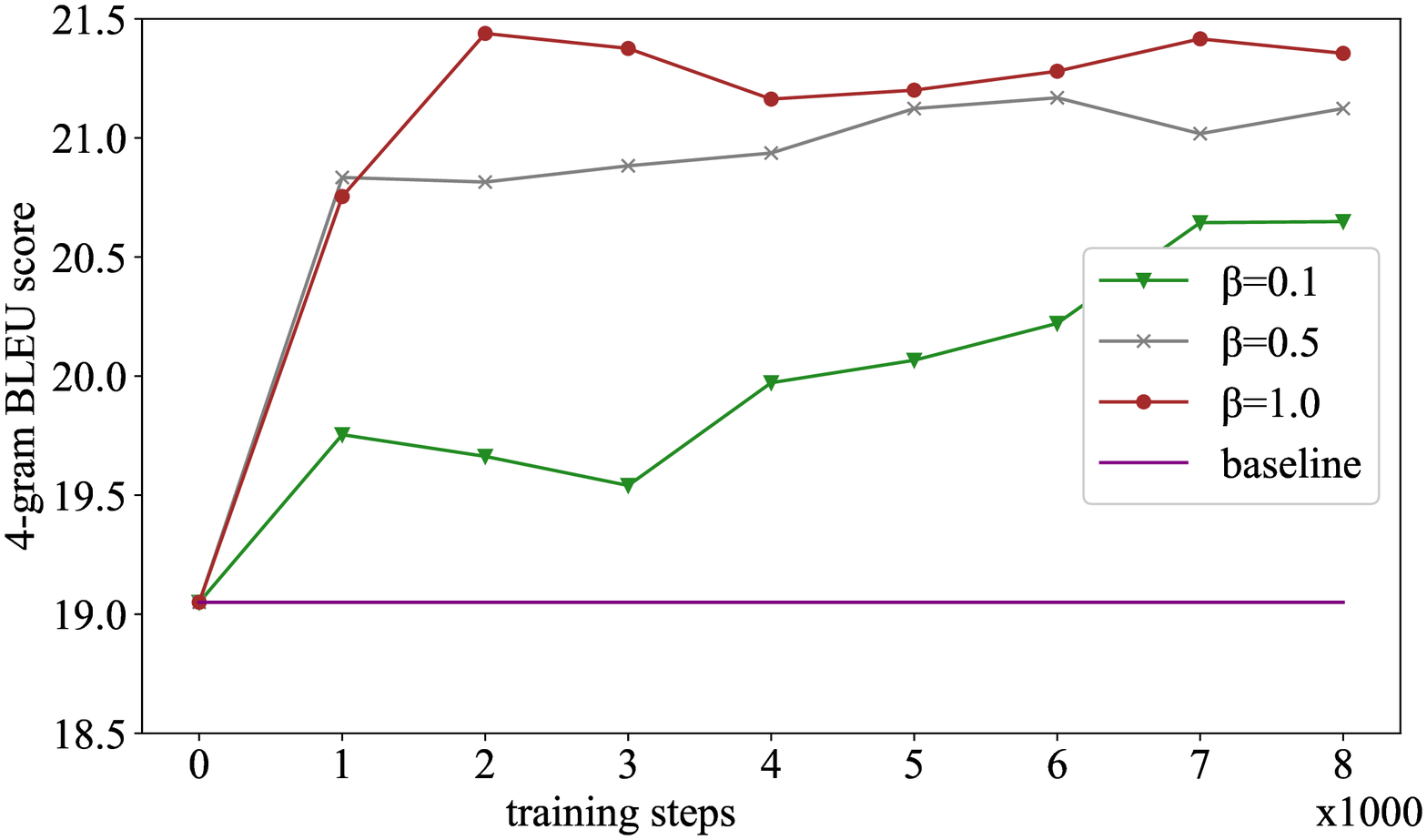}
  \captionsetup{width=.85\linewidth}
  \caption{BLEU scores on tst2010, using ASR transcripts as inputs, varying $\beta$}
  \label{fig:decoder_asr_beta}
\end{minipage}\hfill
\begin{minipage}{0.5\textwidth}
  \centering
  \includegraphics[width=0.85\textwidth]{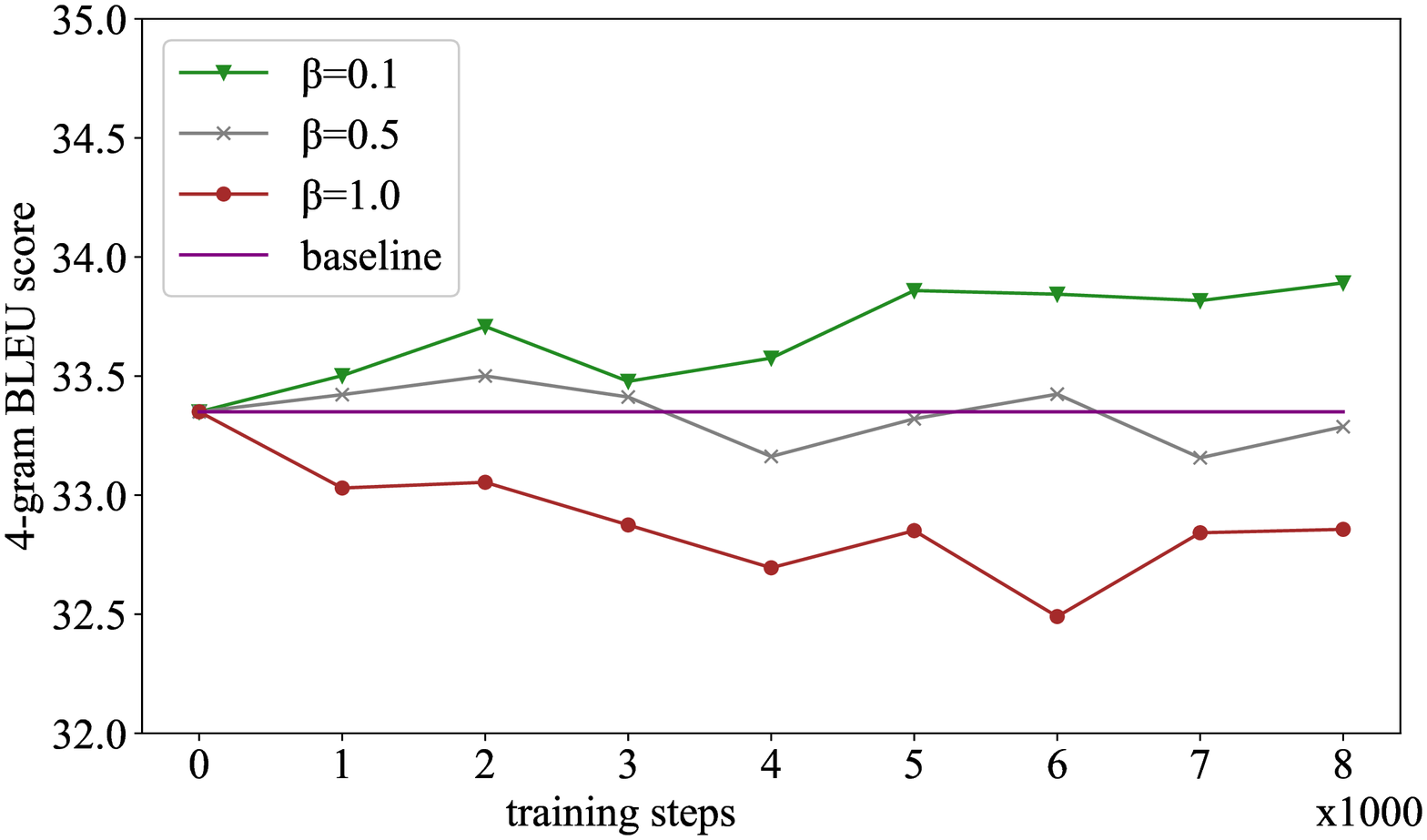}
  \captionsetup{width=.85\linewidth}
  \caption{BLEU scores on tst2010, using clean reference transcripts as inputs, varying $\beta$}
  \label{fig:decoder_ref_beta}
\end{minipage}
\end{figure*}

\subsection{Results}

Firstly, experiments on $\mathcal{L}_{enc}$ and $\mathcal{L}_{dec}$ are performed individually. For $\mathcal{L}_{enc}$, we set $\beta$ to 0.0 and varied $\alpha$ in this part. Figure~\ref{fig:encoder_asr_alpha} reports the BLEU scores of speech translation on tst2010 set at different training steps. The training starts from the baseline model with the BLEU score of 19.05. Performance on $\alpha=\{0.1, 0.5, 1.0\}$ are all better than the baseline system, showing that $\mathcal{L}_{enc}$ is effective in improving translation performance. We found that the larger $\alpha$ achieved better scores than the smaller ones, though this improvement is not significant. Meanwhile, the curve shows that a larger $\alpha$ will make the training more unstable. Figure~\ref{fig:encoder_ref_alpha} shows the BLEU scores on tst2010 set with manual transcriptions as input. It is meant to test how adversarial stability training affects the model performance on the clean text. The figure shows that introducing $\mathcal{L}_{enc}$ has some negative impact on the translation of the clean text at the start of the training phrase, but this negative impact will gradually decrease as the training progress goes on. If we choose a proper $\alpha$, say $\alpha=0.1$, we can improve the quality of speech translation while maintaining close performance on the clean text.

For $\mathcal{L}_{dec}$, we set $\alpha$ to 0.0 and varied $\beta$ in this part. Like the previous two figures, Figure~\ref{fig:decoder_asr_beta} and Figure~\ref{fig:decoder_ref_beta} report the BLEU scores of speech translation on tst2010 with the automatic transcription and manual transcription as inputs, respectively. Figure~\ref{fig:decoder_asr_beta} indicates that the data augmentation method we applied is advantageous to speech translation. It makes a huge improvement from the baseline model. Larger $\beta$ has better performance on speech translation. However, as Figure~\ref{fig:decoder_ref_beta} shows, larger $\beta$ may harm the model's performance when fed with clean text. With $\beta$ set to 0.1 or 0.5, we get fair results.

We then incorporated both $\mathcal{L}_{enc}$ and $\mathcal{L}_{dec}$ in the training phase. Table~\ref{tab:combine_two_loss} shows the results. The best performance is obtained with $\alpha = 0.5$ and $\beta = 0.5$, achieving 2.83 BLEU improvement compared with the baseline system.

\begin{table}[ht!]
\centering
\caption{BLEU score on tst2010 for speech translation system}
\label{tab:combine_two_loss}
\resizebox{0.48\textwidth}{!}{
\begin{tabular}{|c|c|c|c|c|c|c|}
\hline
\multicolumn{3}{|c|}{\multirow{2}{*}{System}}                     & \multicolumn{4}{c|}{BLEU score}     \\ \cline{4-7} 
\multicolumn{3}{|c|}{}                                            & dev2010 &  $\Delta$      & tst2010 &   $\Delta$     \\ \hline
\multicolumn{3}{|c|}{Baseline System}                                    & 19.00   &  & 19.05   &  \\ \hline
\multicolumn{3}{|c|}{\cite{cheng2018towards}}                                    & 19.42   & +0.42 & 19.23   & +0.18  \\ \hline
\multirow{4}{*}{Our work} & \multirow{2}{*}{$\alpha =0.1$} & $\beta =0.1$ & 20.75   & +1.75  & 20.67   & +1.62  \\ \cline{3-7} 
                          &                            & $\beta =0.5$ & 21.14   & +2.14  & 21.66   & +2.61  \\ \cline{2-7} 
                          & \multirow{2}{*}{$\alpha =0.5$} & $\beta =0.1$ & 20.89   & +1.89  & 20.75   & +1.70  \\ \cline{3-7} 
                          &                            & $\beta =0.5$ & \textbf{21.25}   & \textbf{+2.25}  & \textbf{21.88}   & \textbf{+2.83}  \\ \hline
\end{tabular}
}
\end{table}

We also carefully implemented the work of \cite{cheng2018towards}, adding Gaussian noise with standard variance equals to 0.01 to the input word embedding, and then trying to enforce consistent outputs with their original counterpart using adversarial stability training. We reported its BLEU score on tst2010 for speech translation task in Table~\ref{tab:combine_two_loss}. It is clear that training with generic artificial noise only brings minor improvement.

\section{Conclusions}

In this paper, we propose a training architecture that uses speech-to-transcription data to robustify an NMT model in a speech translation scenario. As an intermediate objective, we make the encoder produce a similar output through adversarial stability training. We treat the translation of the manual transcription as the reference of automatic transcription to enforce the decoder consistency. This approach allows for easy incorporation of ASR data into an NMT's training process. Experiments on IWSLT2018 speech translation task demonstrate the effectiveness and robustness of the proposed approach.

\bibliographystyle{IEEEtran}
\bibliography{cheng_iwslt2019}

\begin{thebibliography}{10}
\providecommand{\url}[1]{#1}
\csname url@rmstyle\endcsname
\providecommand{\newblock}{\relax}
\providecommand{\bibinfo}[2]{#2}
\providecommand\BIBentrySTDinterwordspacing{\spaceskip=0pt\relax}
\providecommand\BIBentryALTinterwordstretchfactor{4}
\providecommand\BIBentryALTinterwordspacing{\spaceskip=\fontdimen2\font plus
\BIBentryALTinterwordstretchfactor\fontdimen3\font minus
  \fontdimen4\font\relax}
\providecommand\BIBforeignlanguage[2]{{%
\expandafter\ifx\csname l@#1\endcsname\relax
\typeout{** WARNING: IEEEtran.bst: No hyphenation pattern has been}%
\typeout{** loaded for the language `#1'. Using the pattern for}%
\typeout{** the default language instead.}%
\else
\language=\csname l@#1\endcsname
\fi
#2}}

\bibitem{weiss2017sequence}
R.~J. Weiss, J.~Chorowski, N.~Jaitly, Y.~Wu, and Z.~Chen,
  ``Sequence-to-sequence models can directly translate foreign speech,''
  \emph{Proc. Interspeech 2017}, pp. 2625--2629, 2017.

\bibitem{berard2016listen}
A.~B{\'e}rard, O.~Pietquin, L.~Besacier, and C.~Servan, ``Listen and translate:
  A proof of concept for end-to-end speech-to-text translation,'' in \emph{NIPS
  Workshop on end-to-end learning for speech and audio processing}, 2016.

\bibitem{peitz2012spoken}
S.~Peitz, S.~Wiesler, M.~Nu{\ss}baum-Thom, and H.~Ney, ``Spoken language
  translation using automatically transcribed text in training,'' in
  \emph{International Workshop on Spoken Language Translation (IWSLT) 2012},
  2012.

\bibitem{sperber2017neural}
M.~Sperber, G.~Neubig, J.~Niehues, and A.~Waibel, ``Neural lattice-to-sequence
  models for uncertain inputs,'' in \emph{Proceedings of the 2017 Conference on
  Empirical Methods in Natural Language Processing}, 2017, pp. 1380--1389.

\bibitem{post2013improved}
M.~Post, G.~Kumar, A.~Lopez, D.~Karakos, C.~Callison-Burch, and S.~Khudanpur,
  ``Improved speech-to-text translation with the fisher and callhome
  spanish--english speech translation corpus,'' in \emph{Proc. IWSLT}, 2013.

\bibitem{jan2018iwslt}
N.~Jan, R.~Cattoni, S.~Sebastian, M.~Cettolo, M.~Turchi, and M.~Federico, ``The
  iwslt 2018 evaluation campaign,'' in \emph{International Workshop on Spoken
  Language Translation}, 2018, pp. 2--6.

\bibitem{boito2019mass}
M.~Z. Boito, W.~N. Havard, M.~Garnerin, {\'E}.~L. Ferrand, and L.~Besacier,
  ``Mass: A large and clean multilingual corpus of sentence-aligned spoken
  utterances extracted from the bible,'' \emph{arXiv preprint
  arXiv:1907.12895}, 2019.

\bibitem{di2019must}
M.~A. Di~Gangi, R.~Cattoni, L.~Bentivogli, M.~Negri, and M.~Turchi, ``Must-c: a
  multilingual speech translation corpus,'' in \emph{Proceedings of the 2019
  Conference of the North American Chapter of the Association for Computational
  Linguistics: Human Language Technologies, Volume 1 (Long and Short Papers)},
  2019, pp. 2012--2017.

\bibitem{li2018improving}
X.~Li, H.~Xue, W.~Chen, Y.~Liu, Y.~Feng, and Q.~Liu, ``Improving the robustness
  of speech translation,'' \emph{arXiv preprint arXiv:1811.00728}, 2018.

\bibitem{sperber2017toward}
M.~Sperber, J.~Niehues, and A.~Waibel, ``Toward robust neural machine
  translation for noisy input sequences,'' in \emph{International Workshop on
  Spoken Language Translation (IWSLT), Tokyo, Japan}, 2017.

\bibitem{berard2018end}
A.~B{\'e}rard, L.~Besacier, A.~C. Kocabiyikoglu, and O.~Pietquin, ``End-to-end
  automatic speech translation of audiobooks,'' in \emph{2018 IEEE
  International Conference on Acoustics, Speech and Signal Processing
  (ICASSP)}.\hskip 1em plus 0.5em minus 0.4em\relax IEEE, 2018, pp. 6224--6228.

\bibitem{serdyuk2018towards}
D.~Serdyuk, Y.~Wang, C.~Fuegen, A.~Kumar, B.~Liu, and Y.~Bengio, ``Towards
  end-to-end spoken language understanding,'' in \emph{2018 IEEE International
  Conference on Acoustics, Speech and Signal Processing (ICASSP)}.\hskip 1em
  plus 0.5em minus 0.4em\relax IEEE, 2018, pp. 5754--5758.

\bibitem{duong2016attentional}
L.~Duong, A.~Anastasopoulos, D.~Chiang, S.~Bird, and T.~Cohn, ``An attentional
  model for speech translation without transcription,'' in \emph{Proceedings of
  the 2016 Conference of the North American Chapter of the Association for
  Computational Linguistics: Human Language Technologies}, 2016, pp. 949--959.

\bibitem{bansal2017towards}
S.~Bansal, H.~Kamper, A.~Lopez, and S.~Goldwater, ``Towards speech-to-text
  translation without speech recognition,'' \emph{EACL 2017}, p. 474, 2017.

\bibitem{belinkov2017synthetic}
\BIBentryALTinterwordspacing
Y.~Belinkov and Y.~Bisk, ``Synthetic and natural noise both break neural
  machine translation,'' in \emph{International Conference on Learning
  Representations}, 2018. [Online]. Available:
  \url{https://openreview.net/forum?id=BJ8vJebC-}
\BIBentrySTDinterwordspacing

\bibitem{le2017disentangling}
N.-T. Le, B.~Lecouteux, and L.~Besacier, ``Disentangling asr and mt errors in
  speech translation,'' in \emph{MT Summit 2017}, 2017.

\bibitem{ruiz2017assessing}
N.~Ruiz, M.~A. Di~Gangi, N.~Bertoldi, and M.~Federico, ``Assessing the
  tolerance of neural machine translation systems against speech recognition
  errors.'' in \emph{INTERSPEECH}, 2017, pp. 2635--2639.

\bibitem{cho2017nmt}
E.~Cho, J.~Niehues, and A.~Waibel, ``Nmt-based segmentation and punctuation
  insertion for real-time spoken language translation.'' in \emph{INTERSPEECH},
  2017, pp. 2645--2649.

\bibitem{varavs2018restoring}
A.~V{\=a}ravs and A.~Salimbajevs, ``Restoring punctuation and capitalization
  using transformer models,'' in \emph{International Conference on Statistical
  Language and Speech Processing}.\hskip 1em plus 0.5em minus 0.4em\relax
  Springer, 2018, pp. 91--102.

\bibitem{tsvetkov2014augmenting}
Y.~Tsvetkov, F.~Metze, and C.~Dyer, ``Augmenting translation models with
  simulated acoustic confusions for improved spoken language translation,'' in
  \emph{Proceedings of the 14th Conference of the European Chapter of the
  Association for Computational Linguistics}, 2014, pp. 616--625.

\bibitem{liu2018robust}
H.~Liu, M.~Ma, L.~Huang, H.~Xiong, and Z.~He, ``Robust neural machine
  translation with joint textual and phonetic embedding,'' in \emph{Proceedings
  of the 57th Conference of the Association for Computational Linguistics,
  {ACL} 2019, Florence, Italy, July 28- August 2, 2019, Volume 1: Long Papers},
  2019, pp. 3044--3049.

\bibitem{cheng2018towards}
Y.~Cheng, Z.~Tu, F.~Meng, J.~Zhai, and Y.~Liu, ``Towards robust neural machine
  translation,'' in \emph{Proceedings of the 56th Annual Meeting of the
  Association for Computational Linguistics (Volume 1: Long Papers)}, vol.~1,
  2018, pp. 1756--1766.

\bibitem{matusov2005evaluating}
E.~Matusov, G.~Leusch, O.~Bender, and H.~Ney, ``Evaluating machine translation
  output with automatic sentence segmentation,'' in \emph{International
  Workshop on Spoken Language Translation (IWSLT) 2005}, 2005.

\bibitem{goodfellow2014generative}
I.~Goodfellow, J.~Pouget-Abadie, M.~Mirza, B.~Xu, D.~Warde-Farley, S.~Ozair,
  A.~Courville, and Y.~Bengio, ``Generative adversarial nets,'' in
  \emph{Advances in neural information processing systems}, 2014, pp.
  2672--2680.

\bibitem{yu2017seqgan}
L.~Yu, W.~Zhang, J.~Wang, and Y.~Yu, ``Seqgan: Sequence generative adversarial
  nets with policy gradient,'' in \emph{Thirty-First AAAI Conference on
  Artificial Intelligence}, 2017.

\bibitem{papineni2002bleu}
K.~Papineni, S.~Roukos, T.~Ward, and W.-J. Zhu, ``Bleu: a method for automatic
  evaluation of machine translation,'' in \emph{Proceedings of the 40th annual
  meeting on association for computational linguistics}.\hskip 1em plus 0.5em
  minus 0.4em\relax Association for Computational Linguistics, 2002, pp.
  311--318.

\bibitem{sennrich2016neural}
R.~Sennrich, B.~Haddow, and A.~Birch, ``Neural machine translation of rare
  words with subword units,'' in \emph{Proceedings of the 54th Annual Meeting
  of the Association for Computational Linguistics (Volume 1: Long Papers)},
  vol.~1, 2016, pp. 1715--1725.

\bibitem{vaswani2017attention}
A.~Vaswani, N.~Shazeer, N.~Parmar, J.~Uszkoreit, L.~Jones, A.~N. Gomez,
  {\L}.~Kaiser, and I.~Polosukhin, ``Attention is all you need,'' in
  \emph{Advances in Neural Information Processing Systems}, 2017, pp.
  5998--6008.

\bibitem{vaswani2018tensor2tensor}
A.~Vaswani, S.~Bengio, E.~Brevdo, F.~Chollet, A.~Gomez, S.~Gouws, L.~Jones,
  {\L}.~Kaiser, N.~Kalchbrenner, N.~Parmar, \emph{et~al.}, ``Tensor2tensor for
  neural machine translation,'' in \emph{Proceedings of the 13th Conference of
  the Association for Machine Translation in the Americas (Volume 1: Research
  Papers)}, vol.~1, 2018, pp. 193--199.

\end{thebibliography}
\end{document}